%% file: templateArxiv.tex
\documentclass{article}

\usepackage{PRIMEarxiv}

\usepackage[utf8]{inputenc} 
\usepackage[T1]{fontenc}    
\usepackage{hyperref}       
\usepackage{url}            
\usepackage{booktabs}       
\usepackage{amsfonts}       
\usepackage{nicefrac}       
\usepackage{microtype}      
\usepackage{lipsum}
\usepackage{fancyhdr}       
\usepackage{graphicx}       
\graphicspath{{media/}}     
\usepackage{multirow}
\usepackage{subcaption}

\usepackage[acronym]{glossaries}

\makeglossaries
\loadglsentries{./DATA/gls.tex}
\glsenablehyper

\title{Detecting Face Synthesis Using a Concealed Fusion Model
\thanks{ This work was partly supported by The Alan Turing Institute via Bill \& Gates Foundation (INV-001309) \textit{\underline{Citation}}: 
\textbf{In progress ... }} 
}

\author{
  Roberto Leyva, Victor Sanchez, Gregory Epiphaniou,  Carsten Maple \\
  University of Warwick \\
  Coventry, UK\\
}

\begin{document}
\maketitle

\begin{abstract}
Face image synthesis is gaining more attention in computer security due to concerns about its potential negative impacts, including those related to fake biometrics. Hence, building models that can detect the synthesized face images is an important challenge to tackle. In this paper, we propose a fusion-based strategy to detect face image synthesis while providing resiliency to several attacks. The proposed strategy uses a late fusion of the outputs computed by several undisclosed models by relying on random polynomial coefficients and exponents to conceal a new feature space. Unlike existing concealing solutions, our strategy requires no quantization, which helps to preserve the feature space. Our experiments reveal that our strategy achieves state-of-the-art performance while providing protection against poisoning, perturbation, backdoor, and reverse model attacks. 
\end{abstract}

\section{Introduction}

Face image-based identification is widely used in many applications, making it an essential component of authentication systems \cite{2018Hsu}. Face image synthesis poses a problem for many profile-based systems linked to the users' face images, e.g., fake social media accounts and identity fraud.  Moreover, existing presentation attacks, e.g.,  morphological attacks, can be upgraded with the advances of face image synthesis providing the attacker with concealing and extending capabilities as conducting these attacks initially requires real face images. 

Face image synthesis has recently evolved drastically in terms of image quality~\cite{2022irene}. Hence, it is not only important to detect synthesized face images but also to provide detection models with resiliency to common attacks. To that end, we present an effective strategy to protect a model for fake face image detection while providing competitive performance. Our contributions are as follows:

\begin{enumerate}
    \item We present a conceal-features fusion strategy to detect fake face images. 
    \item Our fusion strategy provides resiliency against poisoning, perturbations, backdoor, and reverse model attacks.
\end{enumerate}

The rest of this paper is organized as follows. In Section \ref{sec:soa}, we review the most related works. In Section \ref{sec:method}, we present the proposed strategy. Section \ref{sec:experiments} provides experimental results and Section \ref{sec:conclusion} concludes this paper.

\section{Related Work}\label{sec:soa}

\textbf{Detection of synthesized imaging data:} these methods usually rely on detecting imperfections in any depicted face \cite{2018Khodabakhsh}. Afchar \textit{et al}. \cite{2018Afchar} propose a \acrfull{CNN} based on the InceptionV3 model to detect synthesized videos. Their method requires detecting the location of faces followed by registration, alignment, and scaling. Hsu \textit{et al}. \cite{2018Hsu} propose a \acrfull{GAN}-based detector that requires measuring the contrastive loss given by the \acrshort{GAN} discriminator. Marra \textit{et al}. \cite{2018Marra} inspect a set of well-established generic models for image tasks, e.g. IV3, DenseNet, Xception, to detect synthesized imaging data. Their work reveals that standard architectures are natively structured for this task. Nataraj \textit{et al}. \cite{2019Nataraj} propose detecting synthesized face images via a set of co-occurrence matrices prior to using a \acrshort{CNN}, as such matrices provide a more descriptive input space. 
 Maiabno \textit{et al}. \cite{2022irene} train several existing \acrshort{CNN} backbones to detect the synthesis in several color spaces. Their results show that those architectures are very sensitive to color space. Rossler \textit{et al}. \cite{2019Rossler} propose to perform a series of manipulations to obtain more synthesized faces to train a \acrshort{CNN}. Zhang \textit{et al}. \cite{2019Zhang}, by using a \acrshort{GAN}-based model, propose learning the synthesis process by solving an image-to-image translation problem. Guarnera \textit{et al.} \cite{2020Guarnera} propose an anspectral analysis of different transformations and intensity domains, which increases the input descriptiveness. Analyzing facial landmarks to detect synthesized face images is proposed by Tolosana \textit{et al}. \cite{2021Tolosana}. Their work suggests that separate models that are fused can detect the synthesis process by separately analyzing the face components, e.g., nose and eyes. This methodology is also supported by the fact that some synthesizing methods can only replace parts of a face instead of generating a whole new face. Local and global matching is explored by Favorskaya \cite{20121Favorskaya}, however, their method heavily relies on additional features, e.g., those extracted from the background and any artifacts surrounding the face. Fusing models to detect the synthesized videos are explored by Coccomini \textit{et al}. Their method requires analyzing the faces frame-by-frame by using a \acrshort{CNN} and a Vision Transformer \cite{2020Dosovitskiy}.

\textbf{Protection of models:}
Prior work by Jin \textit{et al}. \cite{2004Jin} protects a  model by quantizing the input samples using the Wavelet transform and random templates. Talreja \textit{et al}. \cite{2017Talreja} encode face and iris features by using the Reed-Solomon encoder. A separate \acrshort{CNN} is used on each source to produce the features to be hashed. Kaur and Khanna \cite{2019HarkeeratKaur} propose to randomly project the input features and perform the detection in an alternative feature space following the random slope method. This idea is further investigated in \cite{2018Kaur} by performing a fusion between random numbers and local features with dimensionality reduction. Early fusion by hashing the product of random templates with biometric features computed by Gabor filters is proposed by \cite{2020Kaur}. Maneet \textit{et al}. \cite{2019Maneet} present several strategies to protect models via multi-biometric sources. The authors provide the basis for processing face, finger, hand, and iris information at the sensor, feature, score, rank, and decision levels. The authors suggest that strongly protected models should be able to provide encryption at the template level with low distortion of the latent feature space.

\section{Proposed Strategy}\label{sec:method}

Existing strategies to protect models \cite{2017Talreja,2018Kaur} usually quantize the input space, which inevitably leads to losing important information. The small details are the cornerstone of the state-of-the-art methods in face image synthesis detection \cite{2018Khodabakhsh}. Fusion strategies previously proposed to this end, e.g.,  \cite{2020Guarnera,2021Tolosana}, do not adjust the prediction according to the model posteriors. Such an adjustment can increase the model's security and detection performance simultaneously. However, knowing a priori which parts of the fusion process can boost the detection capabilities is challenging. We consider these aspects to develop our strategy. Specifically, as depicted in Fig. \ref{fig:pipeline}, our strategy requires a bank of models to perform late fusion. We protect the posteriors of all models before using a Bayesian model. This model gives the final score to decide if the face is synthetic or not. We explain the constituent components of our strategy next.

\input{DATA/FIGS/Pipeline}

\textbf{Model bank:} Following \cite{2021Tolosana}, we process the input samples using several models. However, different from \cite{2021Tolosana}, we perform no region-based analysis. To this end, we pre-train separately $K=6$ models and protect the outputs given by their last layer. Let us denote the output  of model $k$ by $\hat{x}^k$ for the input image $x$ of size $n_x \times n_y$. The $k^{th}$ model then produces the mapping $\mathbb{R}^{N \times n_x\times n_y} \rightarrow \mathbb{R}^{N \times 2}$, for a set of $N$ images and two classes, i.e., real and synthetic. A model bank comprising $K$ models produces the posterior matrix $\hat{X} \in \mathbb{R}^{N \times 2K }$, which we aim to protect. Note that our model bank can comprise any model, including proprietary ones, whose architecture may remain undisclosed \cite{2018Marra}.

\textbf{Late fusion:} Following the random slope method  \cite{2019HarkeeratKaur}, we propose to protect the decisions of the $K$ models in the bank with random-degree polynomials. Because we can generate polynomials with coefficients and exponents, $\lbrace \alpha_i,\beta_i \rbrace \in \mathbb{Z}$, respectively,  we have a fully discrete domain. The proposed fusion is then performed in a discrete rather than a continuous domain, which avoids quantization. Formally,  the $n^{th}$ sample $\hat{x}_n$ of the matrix produced by the model bank, i.e., $\hat{X} = \lbrace \hat{x}_1,\hat{x}_2, \ldots, \hat{x}_n \rbrace$, is mapped by the function $\rho(\cdot)$:

\begin{subequations}
\begin{equation}
\rho(\hat x_n \mathbf Q^k) = \sum_{i=1\ldots{Q^k}} \alpha^k_i \left( \hat x_n \right) ^ {\beta^k_i}    
\end{equation}
\begin{equation}
    \mathbf Q^k = \left\lbrace \alpha^k_1 \ldots \alpha^k_{Q^k}, \beta^k_1 \ldots \beta^k_{Q^k} \right\rbrace ,
\end{equation}
\end{subequations}

\noindent where $\mathbf{Q}^k \in \mathbb{Z} $ is the set of randomly generated integers used for protection which constitute the key. The design matrix $\hat{X}$ is protected by a vector of size:

\begin{equation}
    P = K \sum_k Q_k .
\end{equation}

The mapping of the projected matrix $\hat{X} \rightarrow \hat{X}_\rho$ is then as:

\begin{equation}
    \hat X_\rho = \left\lbrack \rho( \hat X, \mathbf Q^1), \rho( \hat X, \mathbf Q^2), \ldots \rho( \hat X, \mathbf Q^k)
\right\rbrack
\end{equation}

For instance, if we map $\hat{X}$ with 3-degree polynomials using a bank of $K=6$ models,  we have $P$ as a 36-integer set, where each $(\alpha_i,\beta_i)$ requires 8 bits, making the key's length equal to $8 \times P$ bits. Breaking such a long key is highly unfeasible using standard computing. Our strategy then \emph{fuses and conceals}  the posteriors of all models in the bank without quantization. This aspect adds an authentication-level capability to the inference model. Even if the attacker knows the architecture of the fused models, the key is still required to make predictions and inspect the outputs given by the final decision model.

\textbf{Bayesian model:} Bayesian models have been recently shown to be less prone to overfitting and capable of solving sub-parametrization problems \cite{2022Sanae}. We then use this model as a binary classifier to predict whether a face image is real or fake. The input to this classifier is the matrix $\hat{X}_\rho$ produced by the late fusion encoding and encryption, thus each input $\rho(\hat{x}) = \hat{x}_\rho$  has a dimension of $2K$. We use two fully connected (FC) layers to calculate the final score. The Bayesian model requires estimating the set of probabilistic parameters $\theta = \lbrace \mu, \Sigma \rbrace$, i.e., means $\{\mu\}$ and variances $\{\Sigma\}$,  at each FC layer. Let us consider the target variable $t$ from the vector $\hat{x}_\rho$, whose conditional distribution $p(t \vert \hat{x})$ is Gaussian\footnote{For notation simplicity, we use $\hat{x}=\rho(\hat{x})$.}. For a neural network model mapping function $f(\hat{x}, w)$, with parameters $w$, and inverse variance $\Sigma ^{-1}$, we have:

\begin{equation}
    p(t \vert \hat{x}, w, \Sigma) = \mathcal{N} \left( t \vert f(\hat{x},w), \Sigma ^{-1} \right),
\end{equation}

\noindent where $ p(w, \mu) = \mathcal{N}(w \vert 0, \mu ^{-1} \mathbf{I})$. 
For $N$ observations of $\hat{X}$ with target values $\mathcal{D} = \lbrace t_1, t_2 , \ldots t_N \rbrace$, the likelihood function is:

\begin{equation}
    p(\mathcal{D} \vert w, \Sigma) = \prod \limits_{\forall n} \mathcal{N}(t_n \vert f(\hat{x}_n, w), \Sigma ^{-1}).
\end{equation}

The desired posterior distribution is:

\begin{equation}
    p(w \vert \mathcal{D}, \mu, \Sigma) \approx p(w \vert \mu) ~ p(\mathcal{D}\vert w, \Sigma).
\end{equation}

\noindent It can be proved that the parameter set given by the \acrshort{MAP} estimation is \cite{bishop2006pattern}:

\begin{equation}
    p(t \vert \hat{x}, \mathcal{D}, w, \Sigma) = \mathcal{N} \left( t \vert f(\hat{x},w_{\mathrm{MAP}}), \sigma^{2}(\hat{x}) \right),
\end{equation}

\noindent where the input-dependent variance $\sigma$ is given by:

\begin{subequations}

    \begin{equation}
    \sigma^{2}(\hat{x}) = \Sigma^{-1}+g^{\top}(\mu \mathbf{I} + \Sigma ~ \mathbf{H}) ^{-1}g,
\end{equation}

\begin{equation}
g = \nabla_{w} f(\hat{x} \vert w) \big\vert _{w = w_{\mathrm{MAP}}},
   \end{equation} 
\end{subequations}

\noindent where $\mathbf{H}$ is the Hessian matrix comprising the second derivatives of the sum of square errors with respect to the components of $w$. The distribution $p(t \vert \hat{x}, \mathcal{D})$ is a Gaussian distribution whose mean is given by the neural network function $f(\hat{x}, w_{\mathrm{MAP}})$ and maximizes the posterior likelihood. We can then calculate the model posterior confidence as follows:

\begin{equation}
\underset{c}{\mathrm{max}} ~ f(\hat{x}, w_{\mathrm{MAP}} ) = t_c,
\end{equation}

\noindent where $c$ denotes the two classes, i.e., real or fake. Since for each $\hat{x}$, $t_c \gg t_{\neq c} $, we can then use this result to set the model confidence and make predictions.

\section{Experiments}\label{sec:experiments}

\input{DATA/FIGS/FIG_ACC.tex}

We use the \acrshort{FFHQ} \cite{2017KarrasTero} dataset, which comprises 70K real samples, to produce 70K fake samples using three different synthesizers: s\acrshort{GAN}2~\cite{2020KarrasTero}\footnote{\url{https://github.com/NVlabs/stylegan} }, XL-\acrshort{GAN}~\cite{2022Sauer}\footnote{\url{https://github.com/autonomousvision/stylegan-xl}}, and anyCost-\acrshort{GAN}~\cite{2021Lin}\footnote{\url{https://github.com/mit-han-lab/anycost-gan}}. Hence, we have 70K$+$70K$=$140K samples for each synthesizer. The Bayesian model fuses models that are pre-trained separately on each set of 140K images. We train the Bayesian model using a scheduler to detect error plateaus and scale the learning rate accordingly by a power of ten.
 
\textbf{Detection accuracy:} We first evaluate several detection models separately, i.e., with no fusion, in terms of the \acrfull{mAp} for several data splits, where the training and test datasets contain equal proportions of fake and real images. This first experiment allows us to select the $K=6$ models to be used in the model bank of our strategy. Note that one of the evaluated models is a \acrshort{CNN}-\acrfull{MLP} model we propose with pooling and only expanding convolutional filters to capture small artifacts commonly present in synthesized face images, hereinafter called the MLP\_c3n1f3 model (see Appendix A). Fig. \ref{fig:acc}(a) shows \acrshort{mAp} values for several detection models and data splits on the 140K s\acrshort{GAN}2 images. Note that most of the evaluated models require about 30\% of the training data to achieve competitive accuracy on these images. Fig. \ref{fig:acc}(b) shows \acrshort{mAp} values for several detection models on the 140K images of each of the three synthesizers using an 80:20 data split.  We can see that \acrshort{VGG}-19 achieves the best performance. For the case of the 140K anyCost-\acrshort{GAN} images, our \acrshort{MLP}\_c3n1f3 model outperforms \acrshort{VGG}-19. 

\input{DATA/TABS/Results}

Next, we fuse the six best-performing models from the previous experiment using our proposed strategy. Table \ref{table:results} tabulates \acrshort{mAp} values of our strategy (Bayesian fusion) and other state-of-the-art methods, including the best-performing model in the previous experiment. i.e., \acrshort{VGG}-19. The tabulated results are for several data splits and the 140K images of each of the three synthesizers. The proposed strategy attains competitive accuracy even for small data split values. Hence, it requires fewer training samples to perform very well. 

\textbf{Ablation studies:} We analyze the posterior confidence as the loss value declines during the training of the Bayesian model used by our strategy (see Fig. \ref{fig:epochs_n_klengh} (left)). In general, the strategy makes fewer errors when it is more confident. Since we perform a non-linear mapping,  the new feature space may not be as descriptive as the original one. We then evaluate our strategy's accuracy (\acrshort{mAp}) for several key lengths, especially because we observe that the strategy may not converge to a high \acrshort{mAp} value when using long keys. Fig. \ref{fig:epochs_n_klengh} (right) shows the effect of using long keys in terms of the number of training attempts needed for our strategy to converge as the key length increases. We observe that a  key of length 36 easily makes the model converge in the first attempt.

\input{DATA/FIGS/FIG_EpochConfidence}

\textbf{Model attacks:} We measure the success rate of the poisoning, perturbation, reverse, and backdoor attacks as performed on our strategy using the s\acrshort{GAN}2 images. In other words, we measure the success of miss-detecting samples that are correctly detected before the attack.

\input{DATA/FIGS/FIG_attacks.tex}

\underline{Poisoning}: We swap the labels in the training dataset to generate wrong detections by using several infection proportions, i.e.,  the ratio of swapped labels and the total number of samples \cite{2020LiuYunfei}. Fig. \ref{fig:pollute} shows that as the infection proportion increases, the success rate increases but the accuracy during the training decreases. We observe that the attack is most dangerous when 20\%$\sim$30\% of the labels are poisoned. In such a case, the training accuracy and confidence are high and the model is cheated in $\sim$2\% of the testing samples.

\underline{Perturbation}: We corrupt samples by adding noise and blurring them \cite{2021NingRui}. Fig. \ref{fig:perturbation} shows the results as the number of fused models increases.  Despite a very high success rate, the confidence is low compared to the training confidence. Therefore, this attack can be easily detected by inspecting the model's posterior.

\underline{Reverse Model}: This attack is via manipulating the decision layers to make the model fail with specific samples, for example, by feature vector angle deformation or weight surgery without retraining  \cite{2023Zehavi}. We assume the attacker knows some of the fused models. Fig. \ref{fig:reverse} shows that the attacker requires knowledge of the majority of the fused models to succeed, i.e.,  when the \emph{Training confidence} curve,  represented as the dotted red line, is reached. The attack have high success confidence even when only one of the fused models is known.

\underline{Backdoor}: We mark samples using a black patch following \cite{2020HuangShanjiaoyang}, to maliciously change the classification result. Fig. \ref{fig:backdoor} shows how the success confidence increases as more fused models are attacked. Although the success rate remains almost constant, as the number of attacked models increases, it is likely that the strategy miss-classifies the marked samples as intended by the attacker because the success confidence increases. About 5\% of the marked samples are miss-classified as intended when all fused models are attacked, see the \emph{Success Confidence} curve reaching the \emph{Training confidence} curve (red dotted line).

Note that although our strategy is robust by design to poisoning, backdoor, and perturbation attacks, reverse model attacks pose an important threat, which can be mitigated by not disclosing the architecture.

\section{Conclusion}\label{sec:conclusion}

We have proposed a strategy based on fusion to provide concealment of a model trained to detect synthesized face images while simultaneously increasing accuracy when fewer training samples are available. The proposed strategy projects and encrypts the output of the decision layers of several models into a new feature space. Our proposed strategy is simple yet effective and achieves very competitive accuracy. Our findings have the potential to help protect models used for face validation while providing resiliency to common attacks. Future work focuses on cross-dataset evaluations and robustness against more sophisticated attacks, e.g., backdoor injection, adversarial patches, and weight surgery.

\bibliographystyle{unsrt}  
\bibliography{references}

\end{document}

%% file: DATA/gls.tex

\newacronym{LAN}{LAN}{Local Area Networks}

\newacronym{saas}{SaaS}{Software as Service}

\newacronym{nlp}{NLP}{Natural Language Processing}

\newacronym{OSI}{OSI}{Open Systems Interconnection}

\newacronym{POC}{POC}{Proof of Concept}

\newacronym{FRS}{FRS}{Face Recognition Systems}

\newacronym{PRISMA}{PRISMA}{Preferred Reporting Items for Systematic Reviews and Meta-Analyses}

\newacronym{bmm}{BMM}{Bernoulli Mixture Model}

\newacronym{fvbmm}{FV-BMM}{Fisher Vectors - Bernoulli Mixture Model}

\newacronym{GMM}{GMM}{Gaussian Mixture Model}

\newacronym{hmm}{HMM}{Hidden Markov Model}

\newacronym{SVM}{SVM}{Support Vector Machine}

\newacronym{IoC}{IoC}{Indicator of Compromise}	

\newacronym{3dhog}{3DHOG}{3-Dimensional Histogram of Gradients}	

\newacronym{hof}{HOF}{Histogram of Optical Flow}

\newacronym{GAN}{GAN}{Generative Adversarial Network}

\newacronym{SPN}{SPN}{Sensor Pattern Noise}

\newacronym{PCA}{PCA}{Principal Component Analysis}

\newacronym{surf}{SURF}{Speeded Up Robust Features}

\newacronym{esurf}{eSURF}{3D extended Speeded Up Robust Features}

\newacronym{mips}{MIPs}{Motion Interchanges Patterns}

\newacronym{aic}{AIC}{Akaike Information Criterion}

\newacronym{mle}{MLE}{Maximum Likelihood Estimation}

\newacronym{fsmc}{FSMC}{Finite-State Markov Chain}

\newacronym{roi}{ROI}{Region of Interest}

\newacronym{ROC}{ROC}{Receiver Operating Characteristics}

\newacronym{eer}{EER}{Equal Error Rate}

\newacronym{auc}{AUC}{Area Under the Curve}

\newacronym{bof}{BoF}{Bag of Features}

\newacronym{fps}{FPS}{Frames Per Second}

\newacronym{EFDF}{EFDF}{Early Fusion Deep Features}

\newacronym{FC}{FC}{Fully Connected}

\newacronym{ccr}{CCR}{Correct Classification Rate}

\newacronym{OCR}{OCR}{Optical Chacter Recognition}

\newacronym{bdt}{BDT}{Binary Dense Trajectories}

\newacronym{bidt}{BIDT}{Binary Improved Dense Trajectories}

\newacronym{mbh}{MBH}{Motion Boundaries Histograms}

\newacronym{ARA}{ARA}{Architecture Reference Architecture}

\newacronym{em}{EM}{Expectation Maximization}

\newacronym{CDF}{CDF}{Cumulative Distribution Function}

\newacronym{C3D}{C3D \cite{DTran15}}{Convolution 3D}

\newacronym{HMP}{HMP \cite{JAlmeida11}}{Histogram of Motion Patterns}

\newacronym{AE}{AE}{Auto Encoder}

\newacronym{SVR}{SVR}{Support Vector Regression}

\newacronym{LBP}{LBP \cite{MHeikkil09}}{Local Binary Patterns}

\newacronym{ResNet}{ResNet \cite{KHe16}}{Residual Network}

\newacronym{3DResNet}{3DResNet \cite{KHara18}}{3-Dimensional Residual Network}

\newacronym{DenseNet}{DenseNet \cite{GHuang17}}{Dense Network}

\newacronym{ENet}{ENet}{Elastic Network}

\newacronym{CNN}{CNN}{Convolutional Neural Network}

\newacronym{BERT}{BERT}{Bidirectional Encoder Representations from Transformers}

\newacronym{HOG}{HOG \cite{NDalal05}}{Histogram of Oriented Gradients}

\newacronym{IV3}{IV3 \cite{CSzegedy16} }{Inception Version 3}

\newacronym{LSTM}{LSTM}{Long Short Term Memory}

\newacronym{GloVe}{GloVe \cite{JPennington14}}{Global Vectors}

\newacronym{FL}{FL}{Fuzzy Logic}

\newacronym{TSN}{TSN \cite{LWang16}}{Temporal Segment Networks}

\newacronym{DNN}{DNN}{Deep Neural Network}

\newacronym{FV}{FVs}{Fisher Vectors}

\newacronym{MLP}{MLP}{Multi-Layer Perceptron}

\newacronym{kNN}{kNN}{k-Nearest Neighbour}

\newacronym{AMNet}{AMNet \cite{Fajtl18}}{Attention-based Memorability estimation Network}

\newacronym{LASSO}{LASSO \cite{RTibshirani96}}{Least Absolute Shrinkage and Selection Operator}

\newacronym{I3D}{I3D \cite{Carreira17}}{Two-Stream Inflated 3D ConvNet}

\newacronym{GRU}{GRU \cite{KCho14}}{Gate Recurrent Unit}    

\newacronym{FFT}{FFT}{Fast Fourier Transform}    

\newacronym{NIST}{NIST}{National Institute of Standards and Technology}    

\newacronym{FRSLLS}{FRSLLS}{Face Research Lab London Set}

\newacronym{SOTA}{SOTA}{state-of-the-art}

\newacronym{MAP}{MAP}{Maximum A posterior Probability}

\newacronym{mAp}{mAp}{mean Average precision}

\newacronym{VGG}{VGG}{Visual Geometry Group}

\newacronym{FFHQ}{FFHQ}{Flick Faces High Quality}

%% file: DATA/FIGS/Pipeline.tex
\begin{figure*}[t]
\includegraphics[width=\textwidth]{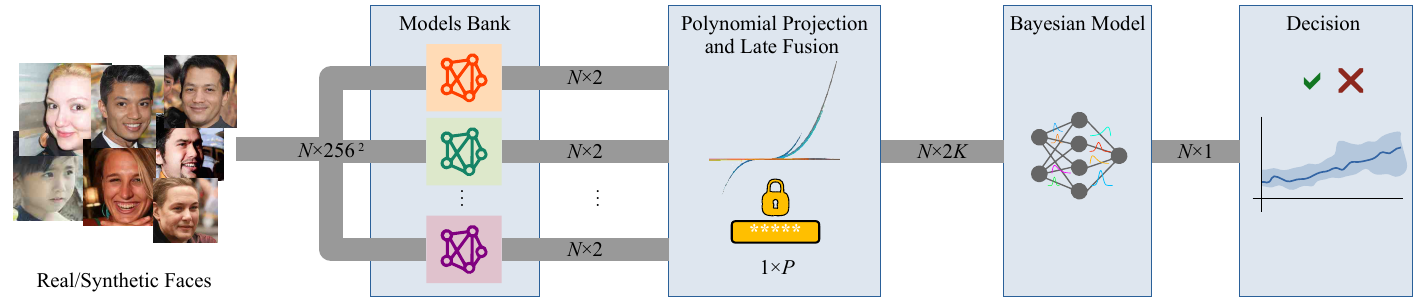}
\vspace{-12pt}
\caption{ \footnotesize{ Our strategy uses a bank of $K$ models. It projects and encrypts the outputs of the decision layer of each individual model to a new feature space. The encrypted projection is used to train a Bayesian model to classify the samples as real or fake.   }} \label{fig:pipeline}
\vspace{-15pt}
\end{figure*}

%% file: DATA/FIGS/FIG_ACC.tex
\begin{figure}[t]
\vspace{-15pt}
\includegraphics[width=\textwidth]{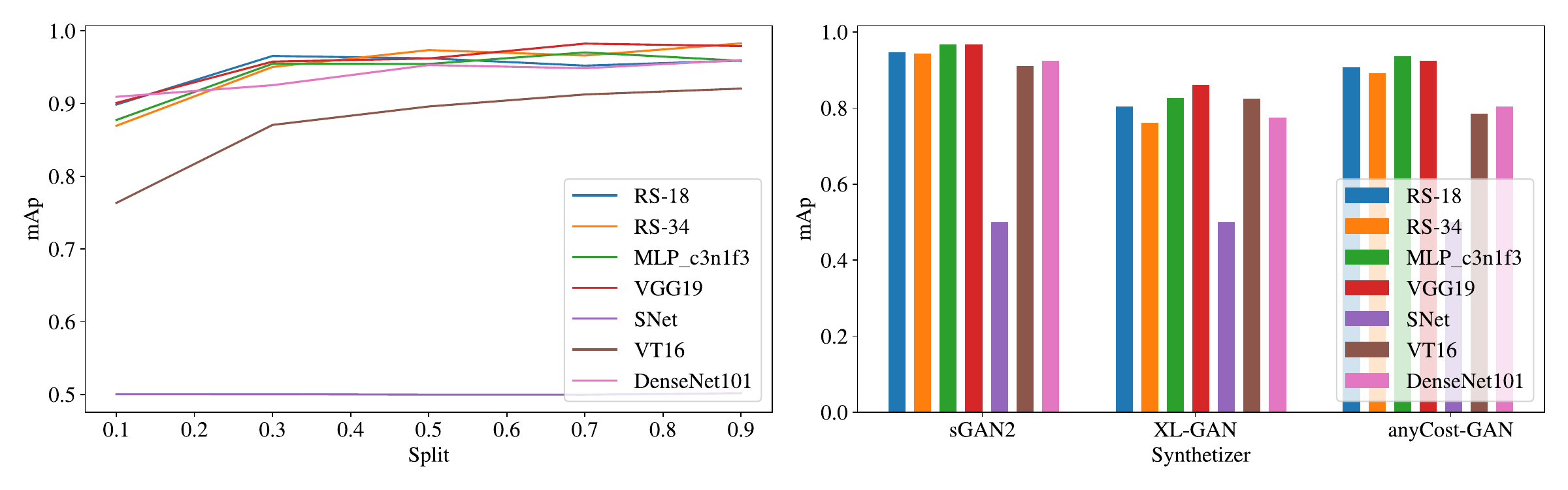}
\vspace{-20pt}
\caption{ \footnotesize{ (left) \acrshort{mAp} values for several data splits using the 140K  sGAN2 images  --  the horizontal axis shows the percentage of training data. (right) \acrshort{mAp} values on the 140K images of each of the three synthesizers using an 80:20 data split. }} \label{fig:acc}
\vspace{-15pt}
\end{figure}

%% file: DATA/TABS/Results.tex
\begin{table}[t]
\centering
\caption{\small{ \acrshort{mAp} values ($\uparrow$) of several detection models on the 140K images of each of the three synthesizers using several data splits.} }
\label{table:results}
\resizebox{1\linewidth}{!}
{

\begin{tabular}{llccccccccc}
\toprule
\multirow{2}{*}{Model} & \multirow{2}{*}{Synthesizer} & \multicolumn{9}{c}{ Data split (proportion of training data)  } \\
&  &  0.1 & 0.2 & 0.3 & 0.4 & 0.5 & 0.6 & 0.7 & 0.8 & 0.9 \\
\midrule
\multirow{3}{*}{\acrshort{VGG}-19\textsuperscript{\dag}} & s\acrshort{GAN}2 \cite{2020KarrasTero} & 0.903 & 0.922 & 0.941 & 0.954 & 0.968 & 0.952 & 0.953 &  0.962 & 0.961 \\
		& XL-\acrshort{GAN} \cite{2022Sauer} & 0.771 & 0.810 & 0.856 & 0.889 & 0.915 & 0.917 & 0.921 &  0.928 & 0.918 \\
  & anyCost-\acrshort{GAN} \cite{2021Lin} & 0.851 & 0.874 & 0.912 & 0.914 & 0.917 & 0.927 & 0.923 &  0.932 & 0.935 \\
  
\multirow{3}{*}{DF~\cite{2022irene}\textsuperscript{\ddag} } & s\acrshort{GAN}2 & 0.875 & 0.895 & 0.913 & 0.931 & 0.935 & 0.942 & 0.946 &  0.945 & 0.949 \\
		& XL-\acrshort{GAN} & 0.802 & 0.856 & 0.899 & 0.901 & 0.908 & 0.912 & 0.917 &  0.916 & 0.915 \\
  & anyCost-\acrshort{GAN} & 0.833 & 0.854 & 0.911 & 0.912 & 0.917 & 0.931 & 0.945 &  0.948 & 0.944 \\

\multirow{3}{*}{CoMat~\cite{2019Nataraj} } & s\acrshort{GAN}2 & 0.901 & 0.934 & 0.956 & 0.954 & \textbf{0.964} & 0.968 & 0.958 &  0.969 & 0.978 \\
		& XL-\acrshort{GAN} & 0.834 & 0.876 & 0.915 & 0.926 & 0.924 & 0.938 & 0.941 &  0.948 & 0.952 \\
  & anyCost-\acrshort{GAN} & 0.831 & 0.884 & 0.932 & \textbf{0.955} & 0.945 & 0.954 & 0.958 &  0.952 & 0.962 \\
  \midrule

\multirow{3}{*}{Bayesian fusion (ours) } & s\acrshort{GAN}2 & \textbf{0.913} & \textbf{0.940} & \textbf{0.964} & 0.954 & 0.951 & \textbf{0.972} & \textbf{0.987} &  \textbf{0.988} & 0.982 \\
		& XL-\acrshort{GAN} & 0.832 & 0.878 & 0.925  & 0.952 & 0.945 & 0.947 & 0.967 &  0.965 & 0.968 \\
  & anyCost-\acrshort{GAN} & 0.871 & 0.910 & 0.951  & 0.961 & 0.944 & 0.957 & 0.968 &  0.974 & \textbf{0.988} \\
\bottomrule
\multicolumn{11}{l}{ \small{\textsuperscript{\dag}Best  performing model based on the first experiment.} \small{\textsuperscript{\ddag}Only compared in the RGB space.} }\\
\\

\end{tabular}
}
\vspace{-32pt}
\end{table}

%% file: DATA/FIGS/FIG_EpochConfidence.tex
\begin{figure}[h]
\centering
\includegraphics[width=1\linewidth]{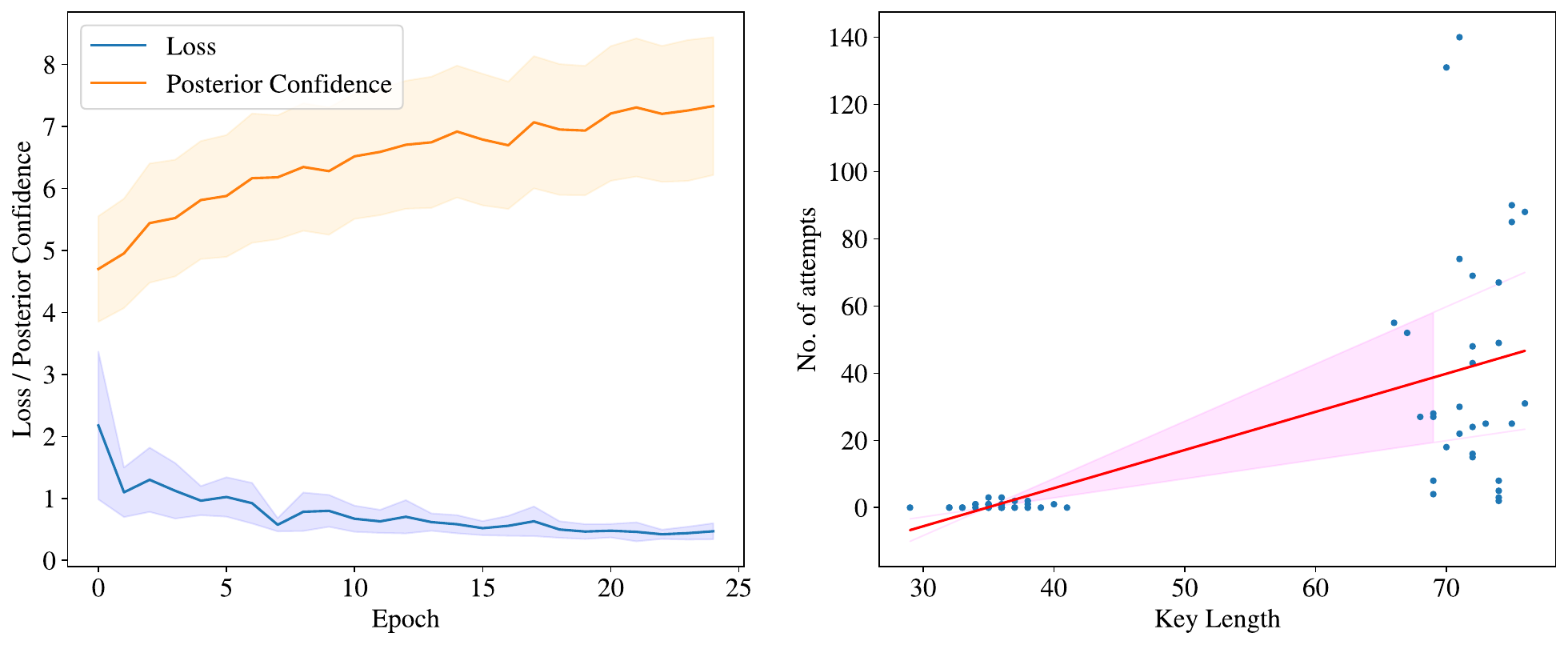}
\vspace{-20pt}
\caption{  \footnotesize{ (left) Loss and posterior confidence values during training. (right) Number of attempts needed for the Bayesian fusion model to converge for several key lengths. }}
\label{fig:epochs_n_klengh}
\vspace{-10pt}
\end{figure}

%% file: DATA/FIGS/FIG_attacks.tex
\begin{figure*}[!h]
\begin{subfigure}{0.48\textwidth}
    \centering
    \includegraphics[width=1\textwidth]{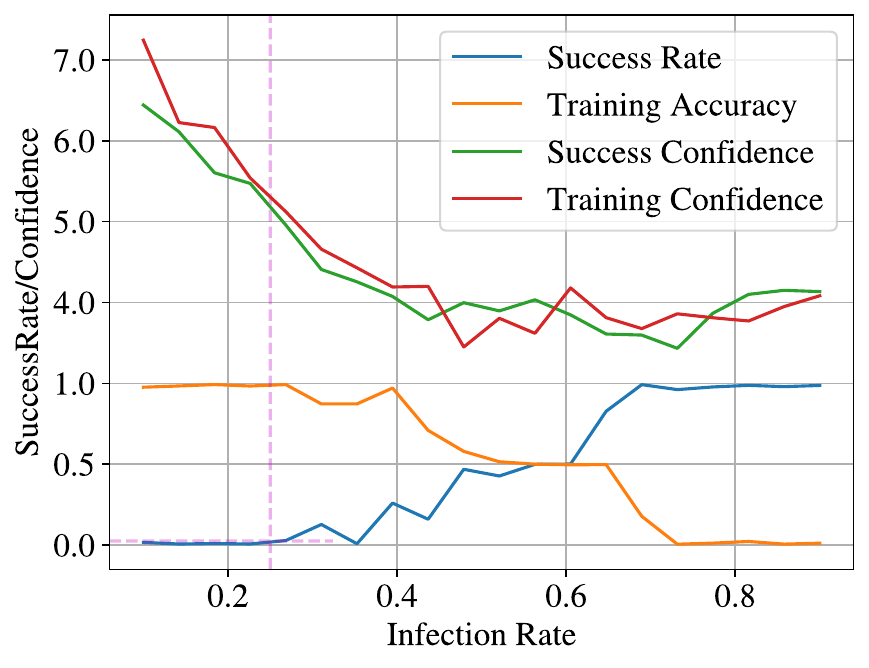}
    \vspace{-15pt}
    \caption{ \scriptsize{Poisoning Attack. }}
    \label{fig:pollute}
\end{subfigure}
\begin{subfigure}{0.48\textwidth}
    \centering
    \includegraphics[width=1\textwidth]{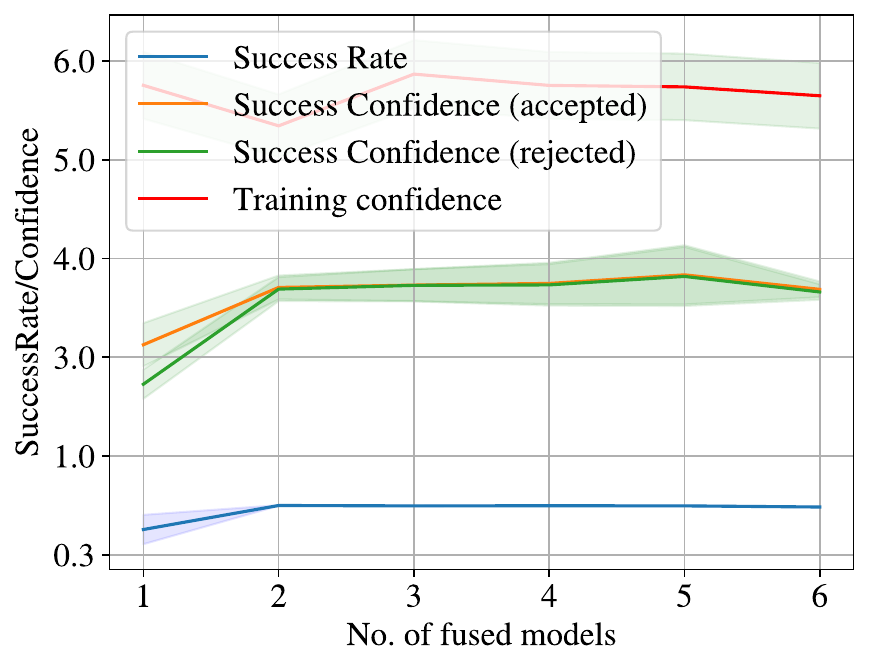}
    \vspace{-15pt}
    \caption{\scriptsize{Perturbation Attack. }}
    \label{fig:perturbation}
\end{subfigure}
\\
\begin{subfigure}{0.48\textwidth}
    \centering
    \includegraphics[width=1\textwidth]{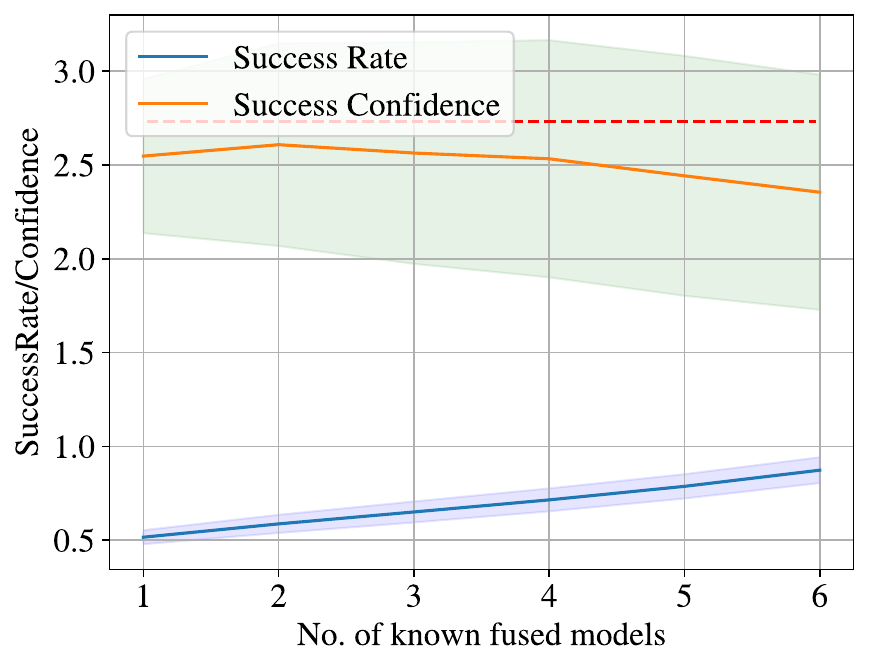}
    \vspace{-15pt}
    \caption{\scriptsize{Reverse Model Attack.} }
    \label{fig:reverse}
\end{subfigure}
\begin{subfigure}{0.47\textwidth}
    \centering
    \includegraphics[width=1\textwidth]{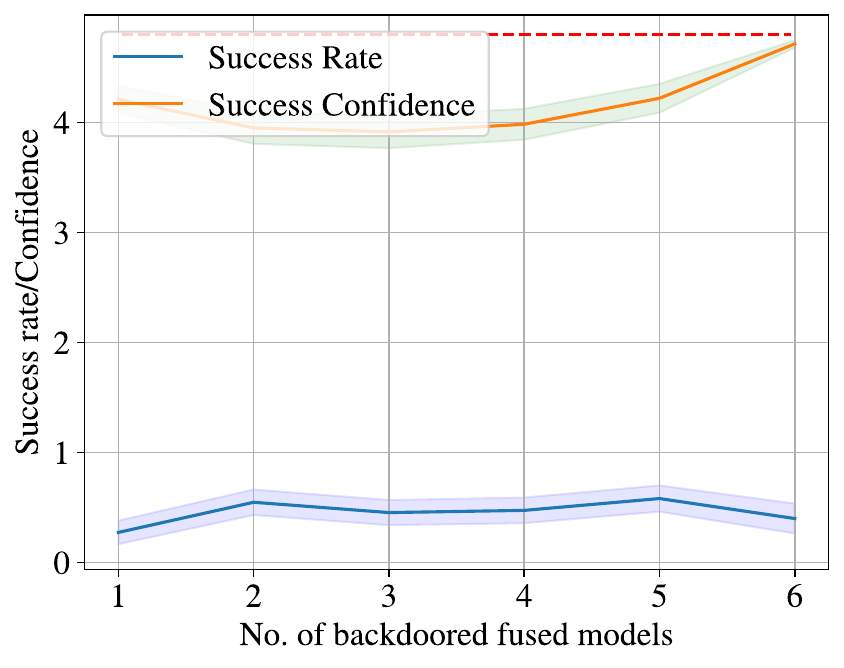}
    \vspace{-15pt}
    \caption{\scriptsize{Backdoor Attack. }}
    \label{fig:backdoor}
\end{subfigure}
\vspace{-5pt}
\label{fig:attacks}
\caption{  \footnotesize{Success rate of several attacks against the proposed Bayesian fusion strategy.}}
\vspace{-18pt}
\end{figure*}